\documentclass{article}

\usepackage{microtype}
\usepackage{graphicx}
\usepackage{booktabs} % for professional tables
\usepackage[nolist,nohyperlinks]{acronym}
\usepackage{subcaption}   % gives \subfloat   (or \subfigure)
\usepackage[skins, breakable]{tcolorbox}
\usepackage{hyperref}

\usepackage[accepted]{icml2025}

\usepackage{amsmath}
\usepackage{amssymb}
\usepackage{mathtools}
\usepackage{amsthm}
\usepackage{enumitem}
\usepackage{algorithm}

\usepackage[capitalize,noabbrev]{cleveref}

\theoremstyle{plain}
\newtheorem{theorem}{Theorem}[section]

\theoremstyle{definition}

\theoremstyle{remark}

\usepackage[textsize=tiny]{todonotes}

\definecolor{AttentionColor}{HTML}{FFE6CC}

\icmltitlerunning{One-Pass to Reason}

\begin{document}

\twocolumn[
\icmltitle{One-Pass to Reason: Token Duplication and Block-Sparse Mask for Efficient Fine-Tuning on Multi-Turn Reasoning}




\begin{icmlauthorlist}
\icmlauthor{Ritesh Goru}{comp}
\icmlauthor{Shanay Mehta}{comp}
\icmlauthor{Prateek Jain}{comp}
\end{icmlauthorlist}

\icmlaffiliation{comp}{DevRev, Palo Alto, United States of America}

\icmlcorrespondingauthor{Ritesh Goru}{ritesh.goru@devrev.ai}
\icmlcorrespondingauthor{Shanay Mehta}{shanay.mehta@devrev.ai}
\icmlcorrespondingauthor{Prateek Jain}{prateek.jain@devrev.ai}

\icmlkeywords{Machine Learning, ICML}

\vskip 0.3in
]



\printAffiliationsAndNotice{} % otherwise use the standard text.

\setcounter{footnote}{1}

\begin{acronym}[GRPO] % Give the longest label here so that the list is nicely aligned
\acro{LLM}{Large Language Model}
\acro{GRPO}{Group Relative Policy Optimisation}
\end{acronym}

\begin{abstract}
Fine-tuning \acp{LLM} on multi-turn reasoning datasets requires N (number of turns) separate forward passes per conversation due to reasoning token visibility constraints, as reasoning tokens for a turn are discarded in subsequent turns. We propose duplicating response tokens along with a custom attention mask to enable single-pass processing of entire conversations. We prove our method produces identical losses to the N-pass approach while reducing time complexity from $O\bigl(N^{3}\bigl)$ to $O\bigl(N^{2}\bigl)$ and maintaining the same memory complexity for a transformer based model. Our approach achieves significant training speedup while preserving accuracy. Our implementation is available online\footnote{\url{https://github.com/devrev/One-Pass-to-Reason}}.
\end{abstract}
\section{Introduction}

Recent progress in \acp{LLM} has sparked a shift from models that directly generate final responses to those that perform explicit intermediate reasoning before generating responses (referred to as reasoning models). Open-source reasoning models, such as DeepSeek-R1 \cite{guo2025deepseek}, demonstrate high performance on several benchmarks. However, these existing reasoning models were trained primarily on single-turn reasoning data.

% Prior works have demonstrated that reasoning significantly improves performance on complex tasks through chain-of-thought (CoT) prompting techniques \cite{Wei2022, lyu-etal-2023-faithful, wang2022self}, which elicit reasoning through carefully designed prompts. Moving beyond prompting. Recent efforts have focused on training models to generate reasoning natively. OpenAI's o1 model demonstrated substantial gains over its non-reasoning predecessor GPT-4o by fine-tuning on reasoning data \cite{jaech2024openai}. However, o1 is proprietary, and we lack knowledge of its training datasets or algorithms. This gap in open research was partially addressed when DeepSeek released DeepSeek-R1 \cite{guo2025deepseek}, an open-source reasoning model that performed comparably to o1, further demonstrating the importance of training models to reason natively.

While numerous studies have investigated fine-tuning LLMs for multi-turn dialogues to improve coherence, context awareness, tool-calling \cite{wang-etal-2025-toolflow, sreedhar-etal-2024-canttalkaboutthis}, these approaches assume non-reasoning dialogues.

Training LLMs for multi-turn reasoning conversations presents novel challenges in managing token visibility. Following industry-standard practices for multi-turn conversations \cite{openaireasoningguide, anthropicreasoningguide}, reasoning models generate internal reasoning tokens, produce a response, and then discard the reasoning tokens from the context in subsequent turns. This creates two fundamental constraints that cannot be addressed with standard multi-turn optimization techniques: (1) \textbf{Visibility Constraints}: Reasoning tokens must be visible during generation but hidden from subsequent conversation turns, requiring conditional visibility that static attention masks cannot satisfy. (2) \textbf{Position ID Discrepancy}: Response tokens follow reasoning tokens during generation but directly follow human messages in a later context, creating positional misalignment.

While prior works have explored masking techniques and position ID assignments to control information flow and enable selective attention within sequences for various pre-training objectives or efficiency gains \cite{wang2024accelerating, du-etal-2022-glm, raffel2020exploring}, none address the specific challenges of multi-turn reasoning conversations where reasoning tokens must be conditionally visible across turns.

This paper addresses these challenges with two primary contributions. (1) We present a theoretical framework featuring a block-sparse visibility mask and strategic position ID assignment scheme that enables processing an entire multi-turn reasoning conversation in a single forward pass while maintaining training correctness (Theorem~\ref{thm:loss_equivalence}). (2) Due to the absence of a publicly available multi-turn reasoning dataset (to the best of our knowledge), we create and release a novel dataset, $\text{MathChat}_{\text{sync}}\text{Reasoning}$, in which each assistant message is augmented with synthetically generated reasoning. (3)  We provide comprehensive empirical validation for the proposed framework on Qwen3 models. % demonstrating that our approach achieves near-linear speedup over naive N-pass baselines ---reducing training complexity from $O(N^3)$ to $O(N^2)$---while preserving model quality across reasoning benchmarks.

\noindent\textbf{Notation.} We use $\mathcal{D}$ to denote a multi-turn reasoning dataset where each conversation $c \in \mathcal{D}$ consists of alternating human messages $h_i$ and assistant messages $a_i$ such that $c = (h_i, a_i)_{i=1}^N$ for $N$ turns. Each assistant message $a_i$ comprises thinking tokens $t_i$ and response tokens $r_i$. We denote $\mathcal{H}_{<i} = (h_j, r_j)_{j=1}^{i-1}$ as conversation history before turn $i$. For token sequence $x$, $s_x$, and $e_x$ represent starting and ending position IDs. The notation $x \rightarrow \mathcal{A}(\cdot)$ indicates sequences that $x$ attends to, and $\mathcal{L}(\cdot)$ denotes language modeling loss (detailed in Appendix~\ref{app:lm-loss}).
\section{Single Pass Fine-tuning on Multi-Turn Reasoning }
\label{problem_formulation}
In this section, we highlight the challenges associated with fine-tuning language models on multi-turn reasoning datasets. We present an optimized approach to process an entire conversation in a single forward pass. In multi-turn reasoning data, response tokens $r_{i}$ must attend to reasoning tokens $t_{i}$ during the generation of $a_{i}$. However, these reasoning tokens must not be visible during subsequent generation of assistant messages $a_{j > i}$. As a result, it is not possible to construct a single static attention mask that supports both conditions in a conversation within a single forward pass—a capability that is often feasible with non-reasoning datasets.

\subsection{N-Pass Approach} 
\begin{figure}[t!]
\vskip 0.2in
\centering
\centerline{\includegraphics[scale=0.36]{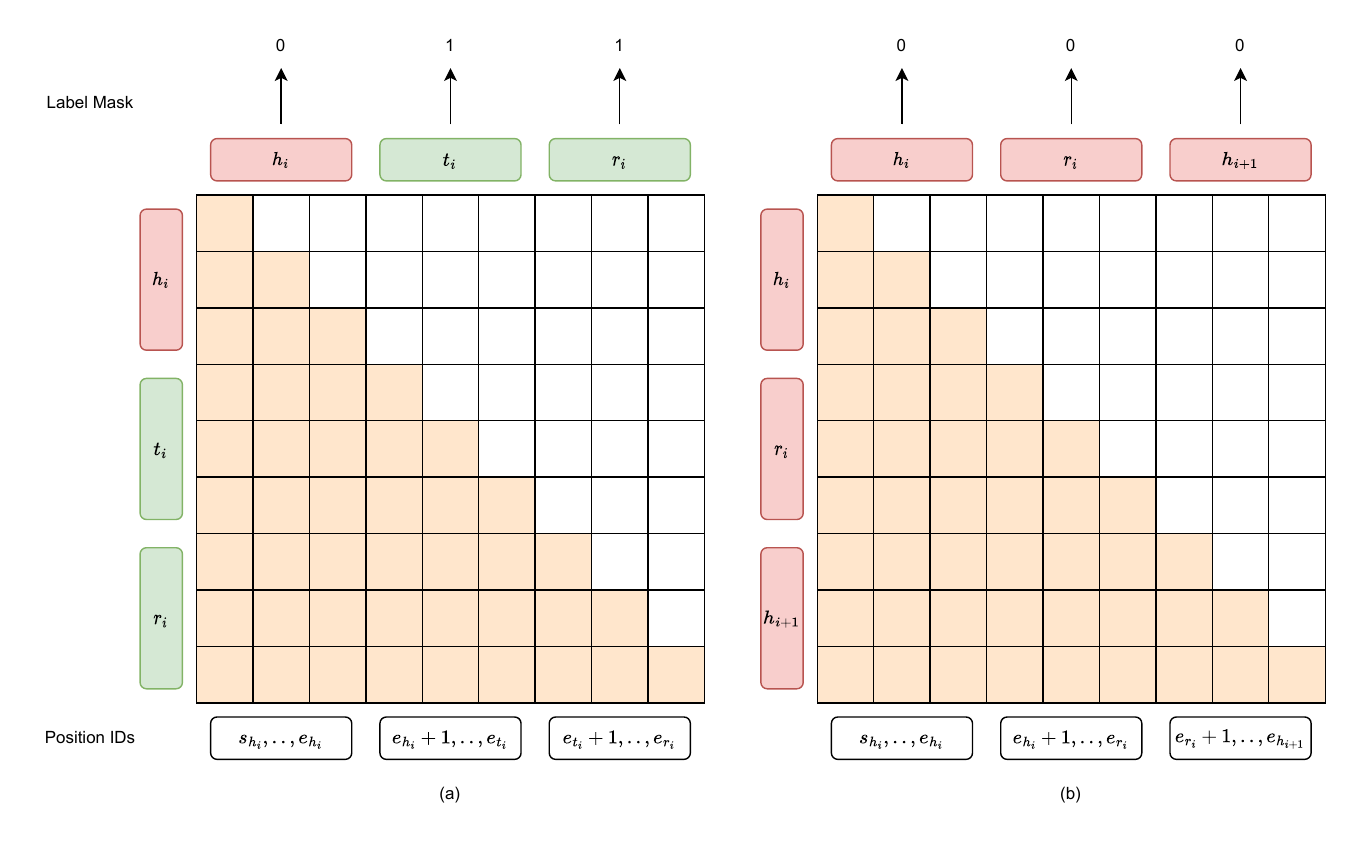}}
    \caption{Causal Attention Masks for N-Pass Approach \fcolorbox{black}{AttentionColor}{\rule{0pt}{2pt}\rule{2pt}{0pt}} represents non-zero attention. (a) Attention Mask for generation of response tokens. (b) Attention Mask when response tokens are in context.}
    \label{fig:causal-mask}
\vskip -0.2in
\end{figure}

A straightforward solution is to perform a separate forward pass for every turn $(\mathcal{H}_{<i}, h_{i}, a_{i})$ of a given conversation $c$. While functionally correct, this approach is computationally inefficient: a conversation with $N$ assistant turns results in $N$ separate training examples. Consequently, the effective size of the dataset increases from $|\mathcal{D}|$ to $|\mathcal{D}| \times N$, inflating training time proportionally. Fig. \ref{fig:causal-mask}(a) shows causal attention mask at the time of generation of $i$th turn response tokens, and Fig. \ref{fig:causal-mask}(b) shows causal attention mask for $i$th turn response tokens when they are part of context during $j>i$ turns.

\subsection{1-Pass Approach}
\label{subsec:cust-mask}
The primary challenge in applying a single forward pass during training due to discrepancy in the attention behavior of $r_{i}$ can be illustrated as follows\footnote{For ease of understanding, we omit the detail that each token within a token sequence also attends to all its preceding tokens, which must be encoded in the attention mask. \label{foot:same_set_tokens}}:
\begin{align*}
    r_{i} \rightarrow \begin{cases}
        \mathcal{A}(\mathcal{H}_{<i}, h_{i}, t_{i}) & \text{generation} \\
        \mathcal{A}(\mathcal{H}_{<i}, h_{i}) & \text{context}
    \end{cases}
\end{align*}
We can resolve this issue through the following steps:
\paragraph{Duplicating response tokens of each assistant message.} We duplicate the response tokens of each assistant message so that one sequence ($r_{i}^{out}$) is used during generation and attends to its associated reasoning tokens. In contrast, the other sequence ($r_{i}^{in}$) is used only as context and does not attend to reasoning tokens.
\paragraph{Custom Attention Mask.} Duplication of response tokens makes it possible to have a single attention mask that satisfies visibility constraints. We define a custom masking strategy for each type of token sequence \( (h_i, t_i, r_i^{\text{in}}, r_i^{\text{out}}) \), ensuring that each token only attends to the appropriate subsequence:
\begin{alignat*}{2}
h_{i} &\rightarrow \mathcal{A}(\mathcal{H}_{<i}^{\textit{in}}) & \qquad r_{i}^{in} &\rightarrow \mathcal{A}(\mathcal{H}_{<i}^{\textit{in}}, h_{i}) \\
t_{i} &\rightarrow \mathcal{A}(\mathcal{H}_{<i}^{\textit{in}}, h_{i}) & \qquad r_{i}^{out} &\rightarrow \mathcal{A}(\mathcal{H}_{<i}^{\textit{in}}, h_{i}, t_{i})
\end{alignat*}

% \begin{figure}[ht]
% \vskip 0.2in
% \begin{center}
% \centerline{\includegraphics[width=\columnwidth]{icml_numpapers}}
% \caption{Historical locations and number of accepted papers for International
% Machine Learning Conferences (ICML 1993 -- ICML 2008) and International
% Workshops on Machine Learning (ML 1988 -- ML 1992). At the time this figure was
% produced, the number of accepted papers for ICML 2008 was unknown and instead
% estimated.}
% \label{icml-historical}
% \end{center}
% \vskip -0.2in
% \end{figure}

\paragraph{Assigning Consistent Position IDs.} After duplication of response tokens, we need to assign consistent position IDs to tokens to maintain the correct relative positions—as if multiple forward passes were performed for each turn in the conversation. If they are assigned sequentially, or the duplicated assistant response tokens share the same position IDs, it will lead to incorrect relative positions. We need a strategic way of assigning position IDs. The following assignment of the first position ID for each token sequence ensures the relative positions are correct and equivalent to N-Pass approach\footnote{\label{posid-note}Position IDs are assigned sequentially based on the order of tokens within each sequence.}:
\[
s_{t_{i}} = s_{r^{in}_{i}} = e_{h_{i}} + 1 \hspace{1.5em} s_{r^{out}_{i}} = e_{t_{i}} + 1 \hspace{1.5em} s_{h_{i+1}} = e_{r^{in}_{i}} + 1
\]
\paragraph{Label Mask.} Duplication of the response tokens also raises the question of which tokens should be included in the loss calculation. The following label mask outlines the inclusion criteria for each token type:
\[
    h_{i} \leftarrow \text{0} \qquad t_{i} \leftarrow \text{1} \qquad r_{i}^{in} \leftarrow \text{0} \qquad r_{i}^{out} \leftarrow \text{1}
\]

\begin{figure}[t!]
\vskip 0.2in
\begin{center}
\centerline{\includegraphics[scale=0.4]{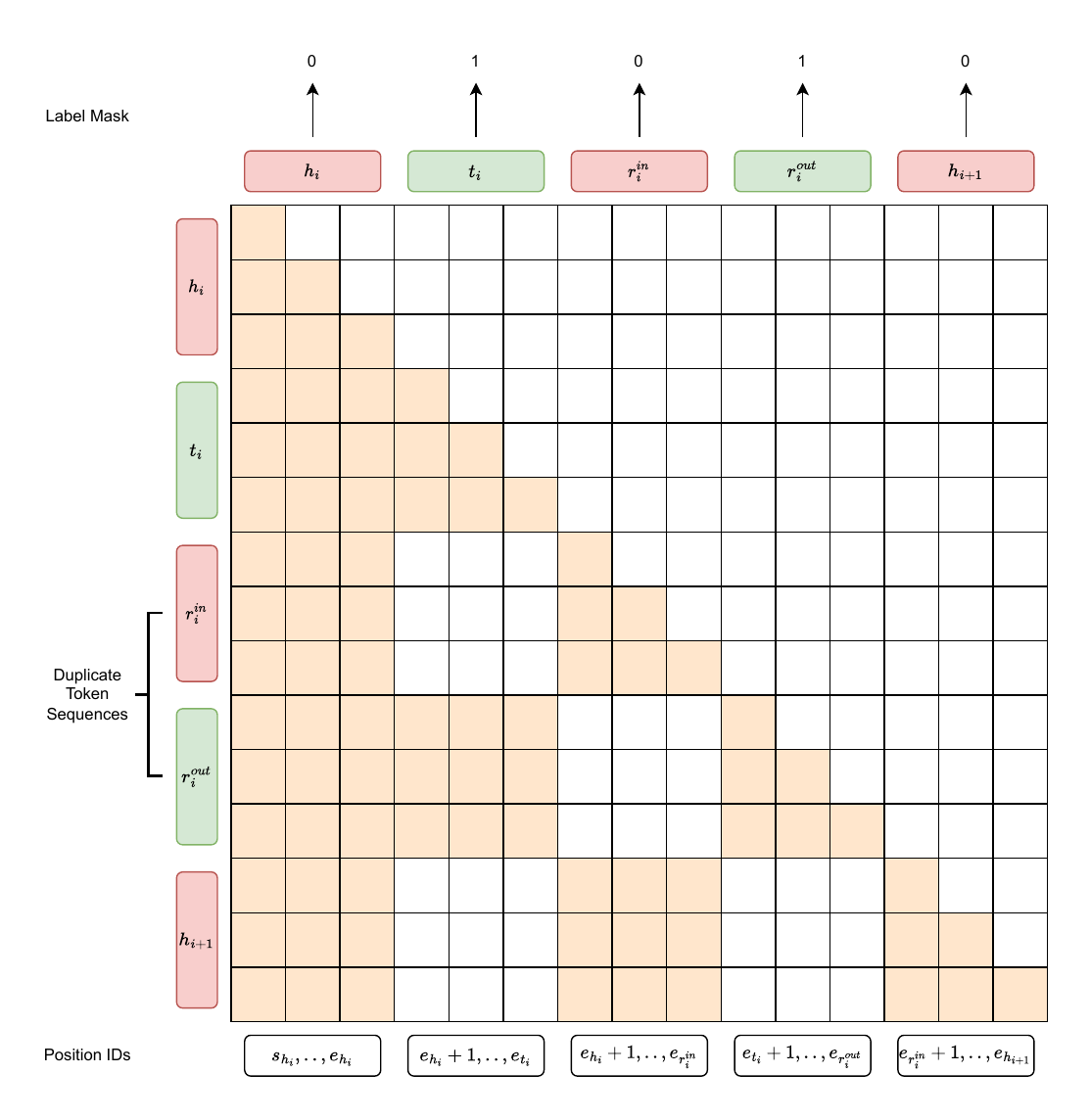}}
\caption{Custom Attention Mask for 1-Pass Approach. \fcolorbox{black}{AttentionColor}{\rule{0pt}{2pt}\rule{2pt}{0pt}} represents non-zero attention.}
\label{fig:cust-mask}
\end{center}
\vskip -0.2in
\end{figure}

Fig. \ref{fig:cust-mask} shows custom attention mask for $i$th turn in the 1-Pass Approach. It combines masks for generation and context from the N-Pass Approach into a single mask with position IDs and a label mask consistent with N-Pass Approach.

\begin{theorem}
\label{thm:loss_equivalence}
    Consider a language model with output distributions determined solely by attention patterns, positional encodings, and input representation. For any conversation \( c \) as input to the model, the sum of the N-Pass language modeling losses is equivalent to the 1-Pass loss:
    \[
    \mathcal{L}^{\textit{1-Pass}}(c) = \sum_{i=1}^{N} \mathcal{L}_{i}^{\textit{N-Pass}}(\mathcal{H}_{<i}, h_{i}, a_{i})
    \]
\end{theorem}

Proof is in the Appendix \ref{app:proof-2.1}

% \begin{figure}[t!]
% \centering
%     \includegraphics[scale=0.32]{figures/attention_mask_custom.pdf}
%     \caption{Custom Attention Mask for Optimized 1-Pass Approach. \fcolorbox{black}{AttentionColor}{\rule{0pt}{2pt}\rule{2pt}{0pt}} represents non-zero attention.}
%     \label{fig:cust-mask}
% \end{figure}

\subsection{Complexity Analysis}
We compare the computational complexity of our 1-Pass method against N-Pass approach for transformer-based models with hidden dimension $d$ \cite{Vaswani2017}. Table~\ref{tab:complexity} summarizes the time and memory complexities for a conversation $c$, where $\ell$ denotes its characteristic turn length.

\begin{table}[h]
\caption{Time and Memory Complexity for N-Pass and 1-Pass Approach}
\label{tab:complexity}
\vskip 0.15in
\begin{center}
\begin{small}
\begin{sc}
\begin{tabular}{lcccr}
\toprule
 & N-Pass & 1-Pass \\
\midrule
$T(c)$    & $O\bigl(N^3 \ell^2 d\bigr)$ & $O\bigl(N^2 \ell^2 d\bigr)$ \\
$M(c)$ & $O\bigl(N^2 \ell^2\bigr)$& $O\bigl(N^2 \ell^2\bigr)$ \\
\bottomrule
\end{tabular}
\end{sc}
\end{small}
\end{center}
\vskip -0.1in
\end{table}

The 1-Pass approach yields an asymptotic time complexity improvement of one order in $N$, offering significant speedups at scale. While it introduces a higher constant memory overhead due to token replication, both methods share the same asymptotic memory complexity. Full derivations are provided in Appendix~\ref{app:complexity}.

\subsection{Efficient Mask Generation}
While our custom attention mask (illustrated in Figure~\ref{fig:cust-mask}) enables single-pass training, generating it involves computing complex visibility patterns across token types and conversation turns. At scale, this computation could become non-trivial, particularly for longer conversations or larger batch sizes. To ensure this remains efficient, we develop an optimized mask generation algorithm that performs all operations on GPU using vectorized tensor operations. Additionally, we simplify the boolean logic for visibility constraints using Karnaugh map reduction, minimizing the number of logical operations required. We provide the complete algorithm in Appendix~\ref{app:mask_generation} for practitioners seeking to implement our method efficiently.
\begin{figure*}[t]
\vskip 0.2in
  \centering
  \subfloat[\scriptsize \parbox{0.8\linewidth}{\centering Speed Analysis across model sizes\\(relative to FA2-N-Pass)}]{
        \includegraphics[width=0.32\linewidth, height=5cm]
            {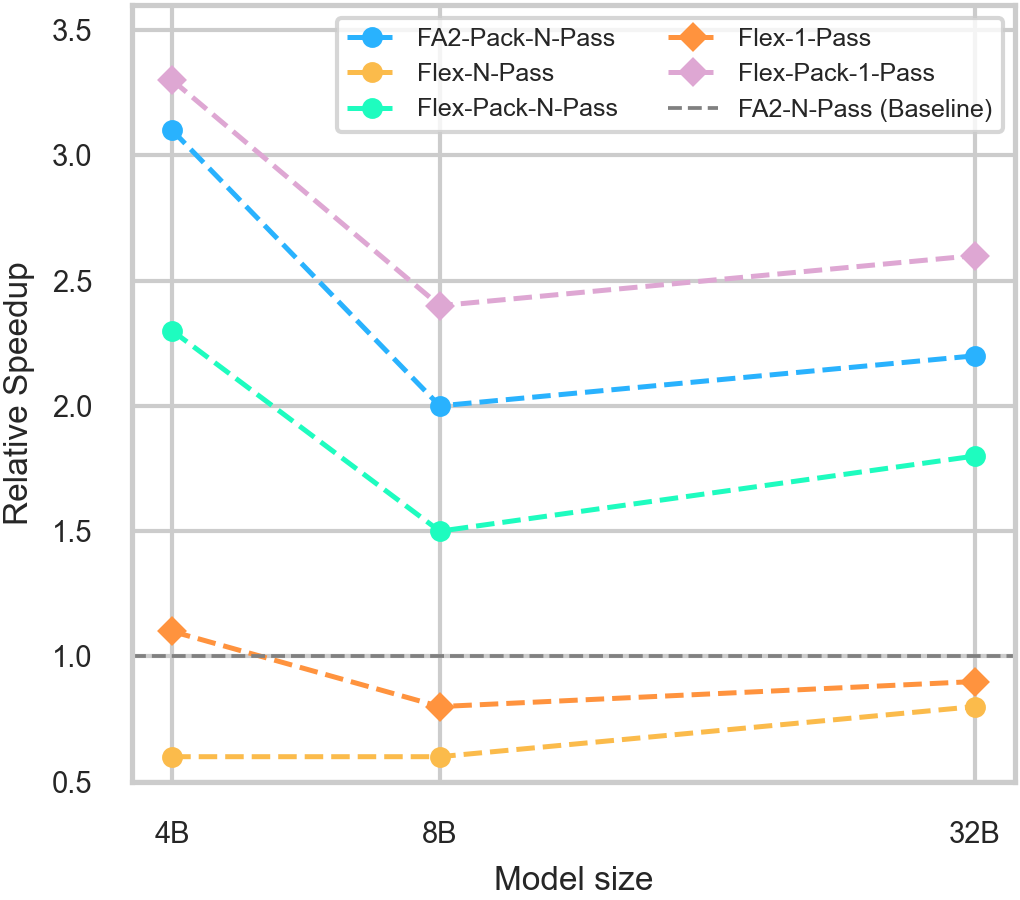}
        \label{fig:speedup-vs-size}}
  \hfill
  \subfloat[\scriptsize \parbox{0.8\linewidth}{\centering K-pass Analysis\\(relative to N-pass)\\Flex-Pack configuration}]{
        \includegraphics[width=0.32\linewidth, height=5cm]
            {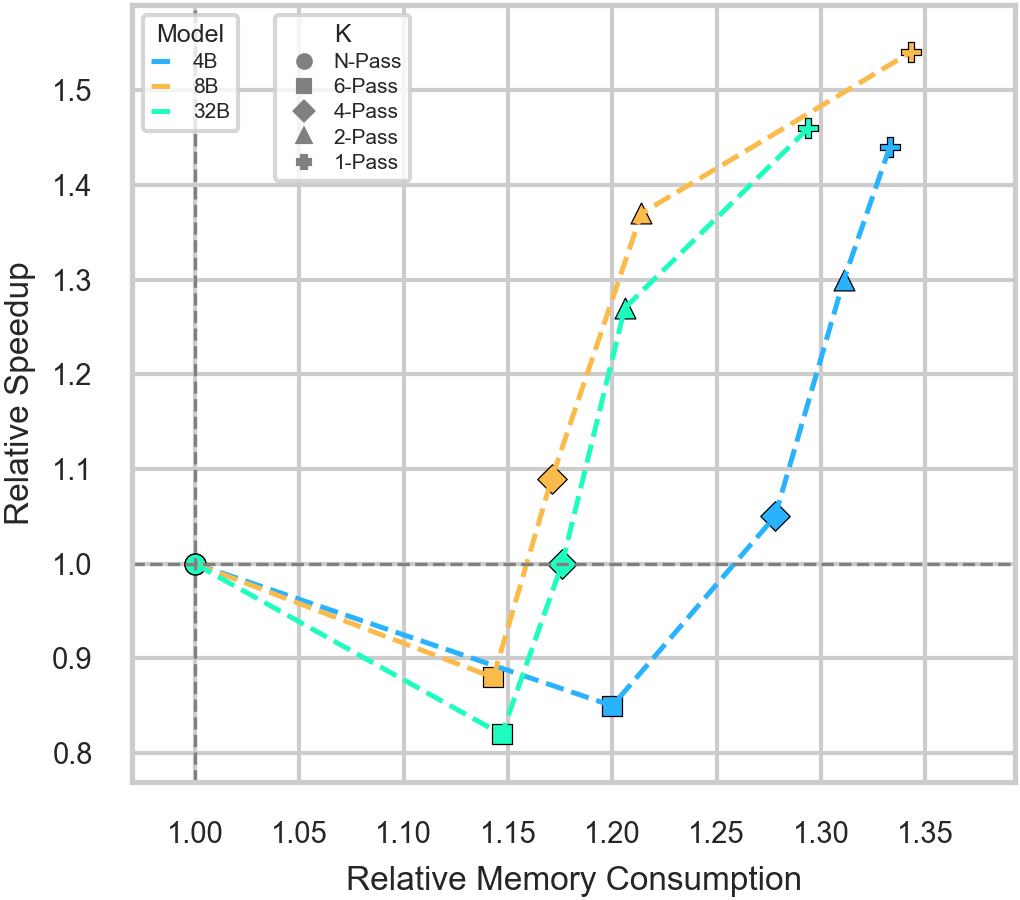}
            \label{fig:kpass-ablation}}
  \hfill
  \subfloat[\scriptsize \parbox{0.7\linewidth}{\centering Conversation Depth Analysis on Qwen3-8B\\(relative to FA2-N-Pass)}]{
        \includegraphics[width=0.32\linewidth, height=5.156cm]
            {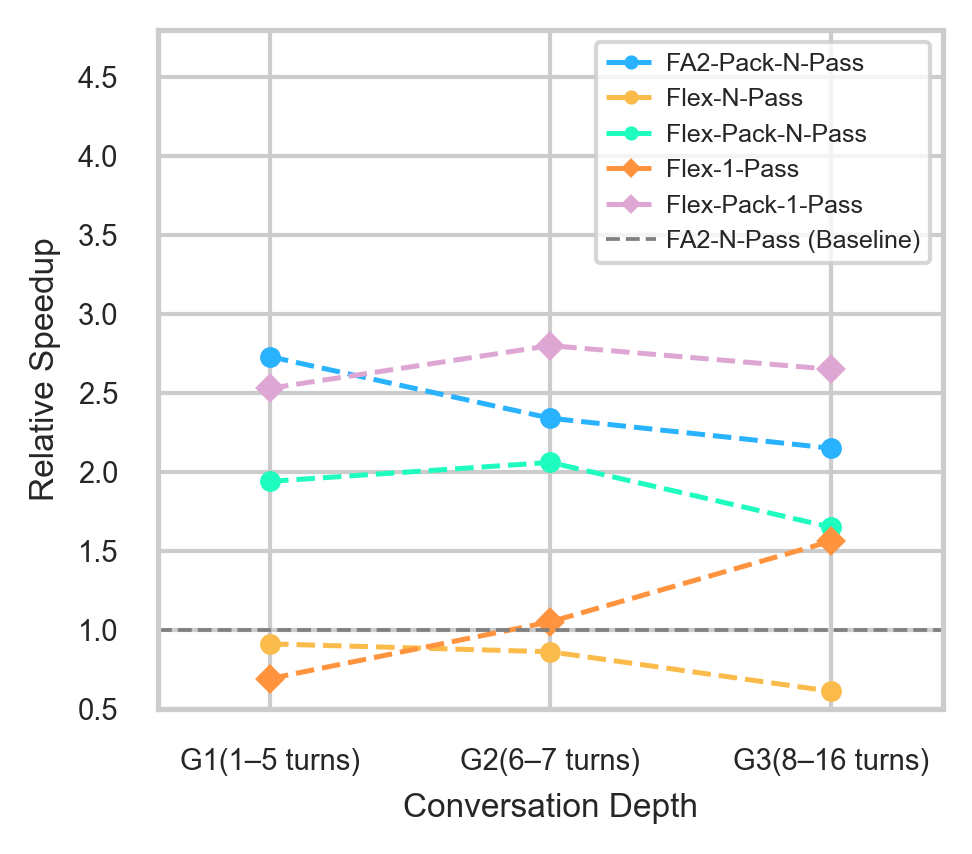}
        \label{fig:conv_depth-ablation}}
  \caption{Training-time experiments}
  \label{fig:all-ablations}
\vskip -0.2in
\end{figure*}

\section{Experiments} \label{sec:experiments}
We evaluate our single-pass fine-tuning on Qwen-3 models (4B, 8B, 32B) with QLoRA~\cite{Dettmers2023}. All experiments were run on a 8$\times$H100 instance (CUDA 12.8, PyTorch 2.7.0), with our method implemented in LLaMA-Factory~\cite{zheng2024llamafactory} and benchmarked against multi-pass baselines. See Appendix \ref{app:expt-setup} for experimental setup.

\subsection{Dataset Creation} \label{sec:dataset}
Addressing lack of a public multi-turn dataset with explicit per-turn reasoning, we introduce $\text{MathChat}_{\text{sync}}\text{Reasoning}$\footnote{\url{https://huggingface.co/datasets/devrev-research/MathChatSync-reasoning}}, derived from $\text{MathChat}_{\text{sync}}$
 ~\cite{liang2024mathchatbenchmarkingmathematicalreasoning}. Assistant turns are augmented with explicit reasoning generated using gpt-4.1-mini, conditioned on dialogue history and current assistant response. Refer to Appendix \ref{app:dataset-creation} for more details. All our experiments are conducted on this dataset.

\subsection{Experimental Setup}
We use FlashAttention2 (FA2)~\cite{dao2024flashattention2} and FlexAttention~\cite{dong2024flexattention} backends. Our 1-Pass method requires a custom attention mask, thus using FlexAttention, as FA2 lacks support for passing custom attention mask; FA2's speed motivates reporting baselines on both for fair comparison. We compare our \textbf{1-Pass method} (with response token duplication) against a standard \textbf{N-Pass baseline} (requiring N forward passes). Both are evaluated with and without sequence packing\footnote{We set the cutoff length to the maximum number of tokens in any datapoint in the dataset for all our experiments.}~\cite{krell2022efficientsequencepackingcrosscontamination}. When packing is enabled, we use llama-factory's \texttt{neat\_packing} implementation: FA2 baselines rely on position IDs to separate packed sequences~\cite{kundu2024enhancing}, while our 1-pass method combines the contamination-free packing mask with our custom attention mask via logical AND.

\subsection{Results:}
\textbf{Training Speedup.} Figure~\ref{fig:speedup-vs-size} shows training speedups. Our 1-Pass method with packing (Flex-Pack-1-Pass) is $1.05\times$, $1.21\times$, and $1.22\times$ faster than FA2-N-Pass baseline with packing (FA2-Pack-N-Pass) on 4B, 8B, and 32B models, respectively. Despite FlexAttention's inherent slowness versus FA2, our method's single-pass efficiency compensates. Compared to N-Pass FlexAttention with packing (Flex-Pack-N-Pass), our Flex-Pack-1-Pass yields $1.44\times, 1.54\times, \text{and } 1.46\times$ speedups for 4B, 8B, and 32B models, respectively. Without packing, our 1-pass method (Flex-1-Pass) lags FA2-N-Pass baseline for 8B and 32B models. We hypothesize that this is because response-token duplication widens the length disparity between conversations, making the method more sensitive to the absence of packing than the N-Pass baseline. Across all experiments, the 1-Pass variants consume roughly 33\% more GPU memory than their N-Pass counterparts.

\textbf{K-Pass Trade-offs.} The 1-Pass and N-Pass approaches represent two extremes: processing the entire conversation in a single pass or in as many passes as there are turns. We therefore also investigate intermediate settings, processing each conversation in K passes. Concretely, we split every dialogue into $K$ contiguous chunks and apply our single-pass mask only to the current chunk, duplicating response tokens and computing loss exclusively for that portion (see Appendix~\ref{app:k-pass} for full details). Figure~\ref{fig:kpass-ablation} reveals a speed-memory trade-off for K$\in${1,2,4,6,N}. Our 1-Pass method maximizes speed with $\sim$33\% more memory (vs. N-Pass). K=2 offers a balance (1.30×--1.37× speedups, $\sim$20\% extra memory). Gains diminish for $K>4$ because, beyond $K=4$, the extra time incurred by the longer sequences created through token duplication outweighs the savings from processing a few turns together.

\textbf{Conversation Scalability.} The dataset contains conversations with depths from 1 to 16 turns. To analyse the effect of depth, we partition it into three groups: G1 (1–5 turns), G2 (6–7 turns), and G3 (8–16 turns)\footnote{This uneven distribution originates from the underlying $\text{MathChat}_{\text{sync}}$ dataset, which is heavily skewed toward 5–7 turn conversations, a bias that propagates to our reasoning corpus.}. Figure~\ref{fig:conv_depth-ablation} shows our Flex-Pack-1-Pass speedups (vs. FA2-Pack-N-Pass) grow with conversation depth ($0.93\times, 1.19\times, 1.23\times$ for G1, G2, G3 respectively). A similar trend appears when comparing our method without packing (Flex-1-Pass) to the FA2-N-Pass baseline: speedups of $0.69\times$, $1.05\times$, and $1.56\times$ for G1, G2, and G3, respectively. This supports the theoretical complexity reduction from $O\bigl(N^{3}\bigl)$ to $O\bigl(N^{2}\bigl)$, as efficiency gains become more pronounced with depth.

These results confirm single-pass training yields significant computational savings, aligning with theoretical advantages, making multi-turn reasoning fine-tuning practical at scale. Please refer Appendix \ref{app:results} for comprehensive results of the experiments conducted.
\section{Conclusion}
We presented an optimized 1-Pass training method for multi-turn reasoning that reduces time complexity from $O(N^3)$ to $O(N^2)$ via strategic token duplication and custom attention mask. Our theoretical analysis confirms loss equivalence with the N-Pass method, enabling efficient training for longer conversations. As multi-turn reasoning becomes central to complex AI tasks, our method offers a scalable and broadly applicable solution. Future work includes exploring adaptive strategies to balance memory-efficiency trade-offs. Additionally, we aim to benchmark performance on latest back-ends such as FlashAttention3 \cite{shah2024flashattention3fastaccurateattention} and port our masking logic to these faster implementations.

\section{Impact Statement}
This paper presents work whose goal is to advance the field of Machine Learning. There are many potential societal consequences of our work, none which we feel must be specifically highlighted here.

\bibliography{example_paper}
\bibliographystyle{icml2025}

%%%%%%%%%%%%%%%%%%%%%%%%%%%%%%%%%%%%%%%%%%%%%%%%%%%%%%%%%%%%%%%%%%%%%%%%%%%%%%%
%%%%%%%%%%%%%%%%%%%%%%%%%%%%%%%%%%%%%%%%%%%%%%%%%%%%%%%%%%%%%%%%%%%%%%%%%%%%%%%
% APPENDIX
%%%%%%%%%%%%%%%%%%%%%%%%%%%%%%%%%%%%%%%%%%%%%%%%%%%%%%%%%%%%%%%%%%%%%%%%%%%%%%%
%%%%%%%%%%%%%%%%%%%%%%%%%%%%%%%%%%%%%%%%%%%%%%%%%%%%%%%%%%%%%%%%%%%%%%%%%%%%%%%
\newpage
\appendix
\onecolumn
\section{Background}
\subsection{Language Modeling Loss} \label{app:lm-loss} For a token sequence ($\mathcal{H}_{<i}$, $h_{i}$, $a_{i}$), the language modeling loss \cite{radford2018improving} for assistant message $a_{i}$ can be expressed as:

\begin{equation}
    \mathcal{L}(\mathcal{H}_{<i}, h_{i}, a_{i}) = - log(P_\Theta(a_{i}|(\mathcal{H}_{<i}, h_{i}))
\end{equation}

where language model is parameterized by $\Theta$.
\section{Proofs}
\subsection{Proof for Theorem 2.1}
\label{app:proof-2.1}
We establish the equivalence by demonstrating that both approaches yield identical probability distributions over sequences, which directly implies equal language modeling losses.

The proof proceeds in three parts: we show that (1) position encodings are equivalent, (2) attention patterns are identical, and (3) the resulting loss functions are mathematically equivalent.

\medskip
\noindent\textbf{Part I: Position Encoding Equivalence.}
Consider the position ID assignments for turn $i$ as defined in Section~\ref{subsec:cust-mask}. In the 1-Pass approach, output tokens receive positions:
\begin{align*}
s_{t_i} &= e_{h_i} + 1 \\
s_{r_i^{out}} &= e_{t_i} + 1
\end{align*}
while input tokens from previous turns $j < i$ receive:
\begin{align*}
s_{r_j^{in}} &= e_{h_j} + 1 \\
s_{h_{j+1}} &= e_{r_j^{in}} + 1
\end{align*}

This assignment ensures that tokens maintain the same relative positional relationships as in the N-Pass approach, where each turn processes tokens sequentially within separate forward passes.

\medskip
\noindent\textbf{Part II: Attention Pattern Preservation.}
The custom attention mask defined in Section~\ref{subsec:cust-mask} ensures causal dependencies are preserved. For turn $i$, the attention patterns are:

\textit{Output tokens:}
\begin{align*}
t_i &\rightarrow \mathcal{A}\left(\mathcal{H}_{<i}^{\textit{in}}, h_i\right) \\
r_i^{out} &\rightarrow \mathcal{A}\left(\mathcal{H}_{<i}^{\textit{in}}, h_i, t_i\right)
\end{align*}

\textit{Input tokens from previous turns $j < i$:}
\begin{align*}
h_j &\rightarrow \mathcal{A}\left(\mathcal{H}_{<j}^{\textit{in}}\right) \\
r_j^{in} &\rightarrow \mathcal{A}\left(\mathcal{H}_{<j}^{\textit{in}}, h_j\right)
\end{align*}

These patterns exactly replicate the causal attention available in the N-Pass approach.

\medskip
\noindent\textbf{Part III: Loss Function Equivalence.}
The language modeling loss for turn $i$ in the N-Pass approach is:
\begin{equation}
\mathcal{L}_{i}^{\textit{N-Pass}}(\mathcal{H}_{<i}, h_i, a_{i}) = -\log P_\theta\left(t_i, r_i \mid \mathcal{H}_{<i}, h_i\right) \label{eq:loss_naive_turn}
\end{equation}

By the autoregressive factorization:
\begin{equation}
\mathcal{L}_{i}^{\textit{N-Pass}}(\mathcal{H}_{<i}, h_i, a_{i}) = -\log P_\theta\left(t_i \mid \mathcal{H}_{<i}, h_i\right) - \log P_\theta\left(r_i \mid \mathcal{H}_{<i}, h_i, t_i\right) \label{eq:loss_naive_expanded}
\end{equation}

The total loss across all turns is:
\begin{equation}
\mathcal{L}^{\textit{N-Pass}}(c) = \sum_{i=1}^{N} \mathcal{L}_{i}^{\textit{N-Pass}}(\mathcal{H}_{<i}, h_i, a_{i}) \label{eq:loss_naive_total}
\end{equation}

For the 1-Pass approach, the loss is computed as:
\begin{equation}
\mathcal{L}^{\textit{1-Pass}}(c) = -\sum_{i=1}^{N} \left[\log P_\theta\left(t_i \mid \mathcal{H}_{<i}^{\textit{in}}, h_i\right) + \log P_\theta\left(r_i^{\textit{out}} \mid \mathcal{H}_{<i}^{\textit{in}}, h_i, t_i\right)\right] \label{eq:loss_opt}
\end{equation}

\textit{Key insight:} Since $r_j = r_j^{\textit{in}} = r_j^{\textit{out}}$ (identical content in different positions) and the position encodings and attention patterns are equivalent as established in Parts I and II, the internal representations are identical. Therefore:
\begin{align}
P_\theta\left(t_i \mid \mathcal{H}_{<i}, h_i\right) &= P_\theta\left(t_i \mid \mathcal{H}_{<i}^{\textit{in}}, h_i\right) \label{eq:prob_equiv_tool}\\
P_\theta\left(r_i \mid \mathcal{H}_{<i}, h_i, t_i\right) &= P_\theta\left(r_i^{\textit{out}} \mid \mathcal{H}_{<i}^{\textit{in}}, h_i, t_i\right) \label{eq:prob_equiv_resp}
\end{align}

Combining equations~\eqref{eq:loss_naive_total}, \eqref{eq:loss_opt}, \eqref{eq:prob_equiv_tool}, and \eqref{eq:prob_equiv_resp}:
\begin{equation}
\mathcal{L}^{\textit{N-Pass}}(c) = \mathcal{L}^{\textit{1-Pass}}(c) \label{eq:loss_equivalence}
\end{equation}
\section{Complexity Analysis}
\label{app:complexity}
\subsection{Input Length}
\subsubsection{N-Pass Approach}
In the N-Pass approach, each turn \(i\) is processed in a separate forward pass. The input to the model at turn \(i\) is:
\[
  \mathcal{H}_{<i},h_i, t_i, r_i
\]
because human and assistant response tokens from previous turns remain in the conversation history, while earlier reasoning tokens are discarded.

Let \(L_{\textit{N-Pass}}\) denote the maximum input length possible for the N-Pass approach for a conversation $c$. It can be defined by:
\begin{equation}
    L_{\textit{N-Pass}} = \sum_{i=1}^{N} (\lvert h_i \rvert + \lvert r_i \rvert) + max_{i=1}^{N}\lvert t_{i} \rvert,
    \label{eq:maxL}
\end{equation}
which is sum of all the human messages and response tokens for entire conversation and maximum length of thinking tokens across turns. To simplify further, assume:
\[
    \lvert h_i \rvert, \lvert t_i \rvert, \lvert r_i \rvert \in O(\ell).
\]
where $\ell$ denote the characteristic turn component length, defined as $\ell$ = $P_{95}({|h_{i}|, |t_{i}|, |r_{i}| : i \in [1,N], c \in \mathcal{D}})$, where $P_{95}$ is the 95th percentile operator.
Then:
\begin{equation}
    L_{\textit{N-Pass}} \in O\bigl((2N + 1) \ell \bigr) = O(N\ell).
\end{equation}

\subsubsection{1-Pass Approach}

Our 1-Pass approach processes the entire conversation \(c\) in a single forward pass. The input length $L_{1-Pass}$ can be calculated as:
\begin{equation}
    L_{\textit{1-Pass}}
    = \sum_{i=1}^{N} \bigl(\lvert h_i \rvert + \lvert t_i \rvert + 2 \lvert r_i \rvert \bigr)
    \in O\bigl(4N\ell\bigr)
    = O\bigl(N\ell\bigr).
    \label{eq:maxLprop}
\end{equation}

\subsection{Time Complexity Analysis}
\label{sec:timecomplexityanalysis}
For a transformer with hidden dimension $d$ and context length $n$, each layer requires $O(n^2 d)$ operations when $n \gg d$~\cite{Vaswani2017}.

\paragraph{N-Pass Approach:} Under the N-Pass approach, each of the \(N\) turns requires a forward pass, each operating on \(O(L_{\textit{N-Pass}}) = O(N\ell)\) tokens. Thus, for conversation \(c\):
\begin{equation}
  T_{\textit{N-Pass}}(c)
  \in O\bigl(N\times(N\ell)^2 d\bigr)
  = O\bigl(N^3 \ell^2 d\bigr).
\end{equation}

\paragraph{1-Pass Approach:} In the 1-Pass approach, all the conversation tokens are given as input at once, thus operating on \(L_{\textit{1-Pass}}\) tokens yielding a cost of:
\begin{equation}
  T_{\textit{1-Pass}}(c)
  \in O\bigl((4N\ell)^2 d\bigr)
  = O\bigl(N^2 \ell^2 d\bigr).
\end{equation}

This represents a factor of $N$ improvement in asymptotic complexity, with substantial gains for large $N$.

\subsection{Memory Complexity Analysis}

A transformer layer with input context length \(n\) has memory complexity \(O(n^2)\) assuming $n \gg d$.

\paragraph{N-Pass Approach:}

 Peak Memory requirement for N-Pass approach is at $L_{\textit{N-Pass}}$ input. Thus for conversation $c$:
\begin{equation}
  \text{M}_{\textit{N-Pass}}(c)
    \in O\bigl((2N+1)^2\ell^2\bigr)
    = O\bigl(N^2 \ell^2\bigr).
\end{equation}

\paragraph{1-Pass Approach:}
 Memory requirement for 1-Pass approach can be given by:
\begin{equation}
  \text{M}_{\textit{1-Pass}}(c)
    \in O\bigl((4N)^2\ell^2\bigr)
    = O\bigl(N^2 \ell^2\bigr).
\end{equation}
Though 1-Pass incurs a higher constant factor due to response token replication, both approaches exhibit identical asymptotic memory complexity.
\section{Experiments}
\subsection{Dataset Creation}
\label{app:dataset-creation}
\begin{figure}[H]
\centering
    \includegraphics[scale=0.25]{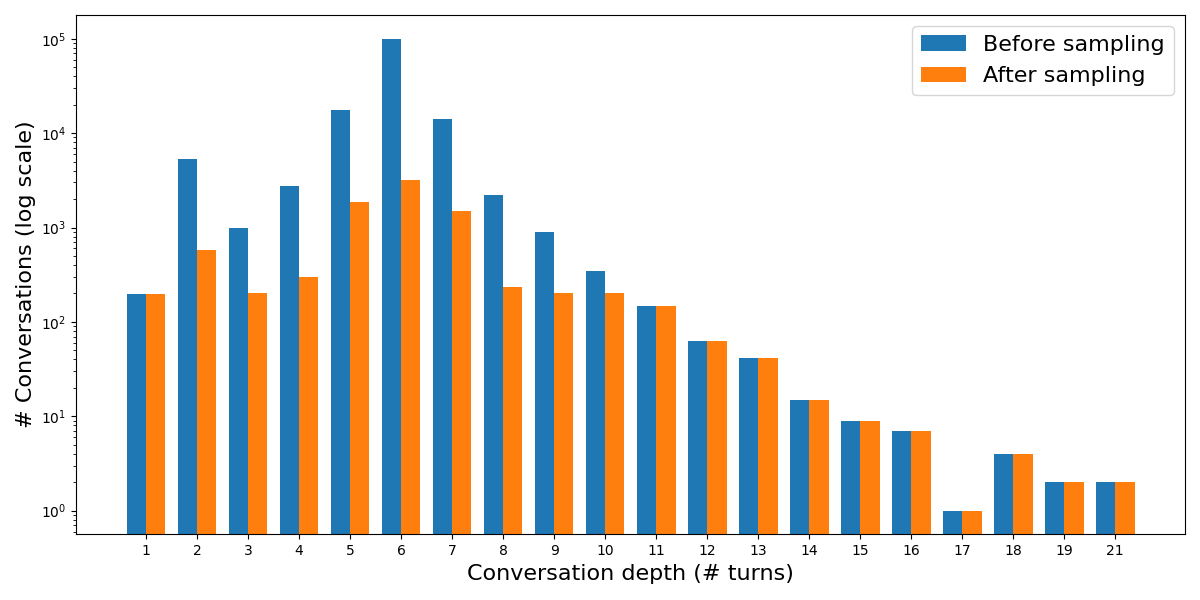}
    \caption{Dataset depth distribution: before vs. after sampling}
    \label{fig:data-dist}
\end{figure}
To enable supervised training with explicit step-by-step reasoning, we construct and release $\text{MathChat}_{\text{sync}}\text{Reasoning}$ along with its generation script. The dataset is obtained by augmenting the original $\text{MathChat}_{\text{sync}}$ corpus~\cite{liang2024mathchatbenchmarkingmathematicalreasoning} with a synthetically-generated rationale for every assistant turn. The procedure comprises three stages.

\paragraph{1. Source corpus.}
$\text{MathChat}_{\text{sync}}$ is a synthetic, dialogue-based mathematics tutoring dataset containing 144,978 conversations with alternating human and assistant messages but no reasoning traces.

\paragraph{2. Depth-balanced sampling.}
Conversation depth in $\text{MathChat}_{\text{sync}}$ is highly skewed toward six-turn dialogues (69 \% of all conversations; see Figure~\ref{fig:data-dist}).
To mitigate this bias, we first down-sample depth-6 dialogues from 100,443 to 30,000 instances.
From the resulting pool we draw a stratified sample of 8,000 conversations.
\begin{itemize}
    \item For each depth $d$, we calculate the proportion of the pool that depth represents.
    \item We allocate to that depth the corresponding proportion of the 8,000-conversation budget, rounding up to the nearest whole conversation.
    \item If the resulting number is below 200, we raise it to (i) 200 or (ii) the total number of conversations available at that depth, whichever is smaller. This guarantees broad coverage across conversation depths.
\end{itemize}
The final split contains 8,797 assistant turns. Figure~\ref{fig:data-dist} compares the depth distribution before and after sampling.
\paragraph{3. Reasoning augmentation.}
For every assistant turn we generate an intermediate reasoning string using gpt-4.1-mini. The model is provided with (i) the dialogue history up to the current human utterance and (ii) the assistant’s reply, and is instructed to output only the hidden rationale that could have produced that reply.
    These rationales are concatenated to the original conversations to form $\text{MathChat}_{\text{sync}}\text{Reasoning}$.

% \begin{table}[ht]
% \centering
% \caption{Conversation–depth distribution in the original MathChat-Sync corpus.}
% \label{tab:depth-before}
% \begin{tabular}{@{}r r@{}}
% \toprule
% Depth & \#Conversations \\
% \midrule
% 1  &   199 \\
% 2  &  5\,339 \\
% 3  &     990 \\
% 4  &  2\,784 \\
% 5  & 17\,507 \\
% 6  &100\,443 \\
% 7  & 13\,964 \\
% 8  &  2\,209 \\
% 9  &     900 \\
% 10 &     349 \\
% 11 &     149 \\
% 12 &      63 \\
% 13 &      41 \\
% 14 &      15 \\
% 15 &       9 \\
% 16 &       7 \\
% 17 &       1 \\
% 18 &       4 \\
% 19 &       2 \\
% 21 &       2 \\
% 23 &       1 \\
% \bottomrule
% \end{tabular}
% \end{table}

\subsection{Efficient mask generation}
\label{app:mask_generation}
We present an efficient algorithm for generating the custom attention mask required by our 1-Pass training method. The algorithm leverages vectorized GPU operations to compute visibility patterns without explicit loops.

\begin{algorithm}[h]
\caption{Efficient Custom Attention Mask Generation}
\label{alg:mask_generation}
\begin{algorithmic}[1]
\REQUIRE Role IDs tensor $\mathbf{R} \in \{0,1,2,3,4\}^{B \times L}$ where $B$ is batch size, $L$ is sequence length
\ENSURE 4D attention mask $\mathbf{M} \in \mathbb{R}^{B \times 1 \times L \times L}$
\STATE \textbf{// Step 1: Compute turn IDs via cumulative sum}
\STATE $\mathbf{R}_{\text{shift}} \leftarrow \text{roll}(\mathbf{R}, \text{shift}=1, \text{dim}=1)$
\STATE $\mathbf{R}_{\text{shift}}[:, 0] \leftarrow 0$
\STATE $\text{turn\_increment} \leftarrow (\mathbf{R} \neq 0) \land (\mathbf{R} = 1) \land (\mathbf{R}_{\text{shift}} \neq 1)$
\STATE $\mathbf{T} \leftarrow \text{cumsum}(\text{turn\_increment}, \text{dim}=1)$
\STATE $\mathbf{T}[\mathbf{R} = 0] \leftarrow 0$ \COMMENT{Zero out padding positions}
\STATE
\STATE \textbf{// Step 2: Create base causal non-padding mask}
\STATE $\mathbf{i} \leftarrow [0, 1, \ldots, L-1]$
\STATE $\text{non\_pad} \leftarrow (\mathbf{R} \neq 0)$
\STATE $\mathbf{M}_{\text{base}} \leftarrow (\mathbf{i}[:, \text{None}] \geq \mathbf{i}[\text{None}, :]) \land \text{non\_pad}[:, :, \text{None}] \land \text{non\_pad}[:, \text{None}, :]$
\STATE
\STATE \textbf{// Step 3: Apply role-specific visibility constraints (K-map optimized)}
\STATE $\text{turn\_equal} \leftarrow (\mathbf{T}[:, :, \text{None}] = \mathbf{T}[:, \text{None}, :])$
\STATE $\mathbf{R}_i \leftarrow \mathbf{R}[:, :, \text{None}]$; $\mathbf{R}_j \leftarrow \mathbf{R}[:, \text{None}, :]$
\STATE $\mathbf{M}_{\text{final}} \leftarrow \mathbf{M}_{\text{base}} \land \big[(\mathbf{R}_j = 1) \lor (\mathbf{R}_j = 4 \land \text{turn\_equal})$
\STATE $\qquad\qquad\qquad\qquad \lor (\mathbf{R}_j = 3 \land \mathbf{R}_i \neq 4) \lor (\mathbf{R}_j = 3 \land \neg\text{turn\_equal})$
\STATE $\qquad\qquad\qquad\qquad \lor (\mathbf{R}_j = 2 \land \text{turn\_equal} \land \mathbf{R}_i \neq 3)\big]$
\STATE
\STATE \textbf{// Step 4: Convert to 4D attention weights}
\STATE $\mathbf{M} \leftarrow \text{where}(\mathbf{M}_{\text{final}}.\text{unsqueeze}(1), 0, -\infty)$
\STATE \textbf{return} $\mathbf{M}$
\end{algorithmic}
\end{algorithm}

\textbf{Implementation Notes:}

\noindent $\bullet$ All operations are performed on GPU using PyTorch's vectorized tensor operations

\noindent $\bullet$ Role IDs: 0 = padding, 1 = human, 2 = thinking, 3 = response (first copy), 4 = response (second copy)

\noindent $\bullet$ The boolean expression in Step 3 is optimized using Karnaugh map reduction to minimize logical operations

\noindent $\bullet$ The algorithm avoids explicit loops by leveraging broadcasting and logical operations

\noindent $\bullet$ For CPU tensors, we temporarily move computation to GPU before returning results to the original device

\subsection{Experimental Setup}
\label{app:expt-setup}
All training runs are initiated using llamafactory-cli in SFT mode. We apply QLoRA with 4-bit NF4 quantization, using a LoRA rank of 32 and a scaling factor of $\alpha = 64$. Training is performed for three epochs with bfloat16 (bf16) precision.

We enable the Liger kernel for improved efficiency. Each GPU processes a batch size of 2, with gradient accumulation over 4 steps. This setup yields an effective batch size of 64 across the 8-GPU node.

\subsection{Comprehensive Results}
\label{app:results}
We report the complete numerical results that support the
figures in Section~\ref{sec:experiments} in Tables \ref{tab:full-throughput}, \ref{tab:kpass} and \ref{tab:depth}.  We report two metrics for
every configuration:

\begin{itemize}[leftmargin=*]
    \item \textbf{Throughput} (``samples per sec.’’) — the average number of
          \emph{full conversations} processed per second.
    \item \textbf{Peak GPU memory} — the peak memory recorded during training.
\end{itemize}
\begin{table}[h]
\footnotesize{}
\begin{tabular}{|l|c|c|c|c|c|}
\hline
\multicolumn{1}{|c|}{Model Size} & Run Setting               & Samples per sec.      & Peak Memory(GB)       & Relative Speedup      & Relative Peak Memory  \\ \hline
\multicolumn{1}{|c|}{4B}         & FA2-N-Pass(Baseline)      & 1.985                 & 9                     & 1.0                   & 1.00                  \\ \hline
                                 & FA2-Pack-N-Pass        & 6.241                 & 9                     & 3.1                   & 1.00                  \\ \hline
                                 & Flex Atten-N-Pass         & 1.286                 & 9                     & 0.6                   & 1.00                  \\ \hline
                                 & Flex Atten+Packing-N-Pass & 4.550                 & 9                     & 2.3                   & 1.00                  \\ \hline
                                 & Flex-1-Pass         & 2.107                 & 12                    & 1.1                   & 1.33                  \\ \hline
                                 & Flex-Pack-1-Pass & 6.552                 & 12                    & 3.3                   & 1.33                  \\ \hline
                                 & \multicolumn{1}{l|}{}     & \multicolumn{1}{l|}{} & \multicolumn{1}{l|}{} & \multicolumn{1}{l|}{} & \multicolumn{1}{l|}{} \\ \hline
\multicolumn{1}{|c|}{8B}         & FA2-N-Pass(Baseline)      & 2.307                 & 14                    & 1.0                   & 1.00                  \\ \hline
                                 & FA2-Pack-N-Pass        & 4.522                 & 14                    & 2.0                   & 1.00                  \\ \hline
                                 & Flex-N-Pass         & 1.365                 & 14                    & 0.6                   & 1.00                  \\ \hline
                                 & Flex-Packing-N-Pass & 3.561                 & 14                    & 1.5                   & 1.00                  \\ \hline
                                 & Flex-1-Pass         & 1.736                 & 18.8                  & 0.8                   & 1.34                  \\ \hline
                                 & Flex-Pack-1-Pass & 5.484                 & 18.8                  & 2.4                   & 1.34                  \\ \hline
                                 & \multicolumn{1}{l|}{}     & \multicolumn{1}{l|}{} & \multicolumn{1}{l|}{} & \multicolumn{1}{l|}{} & \multicolumn{1}{l|}{} \\ \hline
\multicolumn{1}{|c|}{32B}        & FA2-N-Pass(Baseline)      & 0.601                 & 34                    & 1.0                   & 1.00                  \\ \hline
                                 & FA2-Pack-N-Pass        & 1.299                 & 34                    & 2.2                   & 1.00                  \\ \hline
                                 & Flex-N-Pass         & 0.465                 & 34                    & 0.8                   & 1.00                  \\ \hline
                                 & Flex-Packing-N-Pass & 1.078                 & 34                    & 1.8                   & 1.00                  \\ \hline
                                 & Flex-1-Pass         & 0.521                 & 44                    & 0.9                   & 1.29                  \\ \hline
                                 & Flex-Pack-1-Pass & 1.578                 & 44                    & 2.6                   & 1.29                  \\ \hline
\end{tabular}
\caption{\textbf{Throughput and peak memory across execution strategies.}
FA2 = FlashAttention 2; Flex = FlexAttention.  
Pack denotes dynamic sequence-packing; “1-Pass’’ is our proposed approach.  
Relative columns are computed with respect to the corresponding
FA2–N-Pass baseline.}
\label{tab:full-throughput}
\end{table}

\begin{table}
\footnotesize{}
\begin{tabular}{|l|c|c|c|c|c|}
\hline
\multicolumn{1}{|c|}{Model Size} & K                     & Samples per sec.      & Peak Memory(GB)       & Relative Speedup      & Relative Peak Memory  \\ \hline
\multicolumn{1}{|c|}{4B}         & N-Pass(baseline)      & 4.55                  & 9                     & 1.00                  & 1.00                  \\ \hline
                                 & 6-Pass                & 3.89                  & 10.8                  & 0.85                  & 1.20                  \\ \hline
                                 & 4-Pass                & 4.76                  & 11.5                  & 1.05                  & 1.28                  \\ \hline
                                 & 2-Pass                & 5.91                  & 11.8                  & 1.30                  & 1.31                  \\ \hline
                                 & 1-Pass                & 6.55                  & 12                    & 1.44                  & 1.33                  \\ \hline
                                 & \multicolumn{1}{l|}{} & \multicolumn{1}{l|}{} & \multicolumn{1}{l|}{} & \multicolumn{1}{l|}{} & \multicolumn{1}{l|}{} \\ \hline
\multicolumn{1}{|c|}{8B}         & N-Pass(baseline)      & 3.56                  & 14                    & 1.00                  & 1.00                  \\ \hline
                                 & 6-Pass                & 3.13                  & 16                    & 0.88                  & 1.14                  \\ \hline
                                 & 4-Pass                & 3.87                  & 16.4                  & 1.09                  & 1.17                  \\ \hline
                                 & 2-Pass                & 4.87                  & 17                    & 1.37                  & 1.21                  \\ \hline
                                 & 1-Pass                & 5.48                  & 18.8                  & 1.54                  & 1.34                  \\ \hline
                                 & \multicolumn{1}{l|}{} & \multicolumn{1}{l|}{} & \multicolumn{1}{l|}{} & \multicolumn{1}{l|}{} & \multicolumn{1}{l|}{} \\ \hline
\multicolumn{1}{|c|}{32B}        & N-Pass(baseline)      & 1.08                  & 34                    & 1.00                  & 1.00                  \\ \hline
                                 & 6-Pass                & 0.88                  & 39                    & 0.82                  & 1.15                  \\ \hline
                                 & 4-Pass                & 1.08                  & 40                    & 1.00                  & 1.18                  \\ \hline
                                 & 2-Pass                & 1.37                  & 41                    & 1.27                  & 1.21                  \\ \hline
                                 & 1-Pass                & 1.58                  & 44                    & 1.46                  & 1.29                  \\ \hline
\end{tabular}
\caption{\textbf{Speed–memory trade-off as a function of $K$.}
Each dialogue is split into $K$ equal-length chunks that are processed
sequentially in a \emph{single} forward/backward pass.
$K\!=\!N$ corresponds to the per-turn baseline, while $K\!=\!1$
is our single-pass method. All experiments use the FlexAttention backend with sequence packing (Flex-Pack), the configuration that achieved the best overall speed in our primary evaluation.}
\label{tab:kpass}
\end{table}

\begin{table}
\footnotesize{}
\begin{tabular}{|l|c|c|c|c|c|}
\hline
                              & Run Setting               & Samples per sec.      & Peak Memory(GB)       & Relative Speedup      & Relative Peak Memory  \\ \hline
\multicolumn{1}{|c|}{Group 1} & FA2-N-Pass(Baseline)      & 2.54                  & 14                    & 1.00                  & 1                     \\ \hline
                              & FA2-Pack-N-Pass        & 6.93                  & 14                    & 2.73                  & 1                     \\ \hline
                              & Flex-N-Pass         & 2.32                  & 14                    & 0.91                  & 1                     \\ \hline
                              & Flex-Packing-N-Pass & 4.94                  & 14                    & 1.94                  & 1                     \\ \hline
                              & Flex-1-Pass         & 1.74                  & 18.8                  & 0.69                  & 1.34                  \\ \hline
                              & Flex-Pack-1-Pass & 6.43                  & 18.8                  & 2.53                  & 1.34                  \\ \hline
                              & \multicolumn{1}{l|}{}     & \multicolumn{1}{l|}{} & \multicolumn{1}{l|}{} & \multicolumn{1}{l|}{} & \multicolumn{1}{l|}{} \\ \hline
\multicolumn{1}{|c|}{Group 2} & FA2-N-Pass(Baseline)      & 1.02                  & 14                    & 1                     & 1                     \\ \hline
                              & FA2-Pack-N-Pass        & 2.39                  & 14                    & 2.34                  & 1                     \\ \hline
                              & Flex-N-Pass         & 0.87                  & 14                    & 0.86                  & 1                     \\ \hline
                              & Flex-Packing-N-Pass & 2.10                  & 14                    & 2.06                  & 1                     \\ \hline
                              & Flex-1-Pass         & 1.07                  & 18.8                  & 1.05                  & 1.34                  \\ \hline
                              & Flex-Pack-1-Pass & 2.86                  & 18.8                  & 2.80                  & 1.34                  \\ \hline
                              & \multicolumn{1}{l|}{}     & \multicolumn{1}{l|}{} & \multicolumn{1}{l|}{} & \multicolumn{1}{l|}{} & \multicolumn{1}{l|}{} \\ \hline
\multicolumn{1}{|c|}{Group 3} & FA2-N-Pass(Baseline)      & 1.06                  & 14                    & 1                     & 1                     \\ \hline
                              & FA2-Pack-N-Pass        & 2.28                  & 14                    & 2.15                  & 1                     \\ \hline
                              & Flex-N-Pass         & 0.65                  & 14                    & 0.61                  & 1                     \\ \hline
                              & Flex-Packing-N-Pass & 1.75                  & 14                    & 1.65                  & 1                     \\ \hline
                              & Flex-1-Pass         & 1.66                  & 18.8                  & 1.56                  & 1.34                  \\ \hline
                              & Flex-Pack-1-Pass & 2.81                  & 18.8                  & 2.65                  & 1.34                  \\ \hline
\end{tabular}
\caption{\textbf{Impact of conversation depth (Qwen-3 8B).}
Group 1 (1–5 turns), Group 2 (6–7 turns), and Group 3 (8–16 turns).
Our 1-Pass approach gains more speed as depth increases, in line
with the theoretical $O(N^{2})$ vs.\ $O(N^{3})$ complexity gap.}
\label{tab:depth}
\end{table}

\subsubsection{Implementing K-Pass Processing}
\label{app:k-pass}
%======================================================================

To obtain the results in Table~\ref{tab:kpass} we extend our
Optimised 1-Pass scheme to an intermediate \emph{$K$-Pass} schedule.
Assume a conversation contains $N$ assistant turns
$(h_{1}, t_{1}, r_{1}),\dots,(h_{N}, t_{N}, r_{N})$.

\begin{enumerate}[leftmargin=*, label=(\alph*)]
\item \textbf{Chunking the dialog.}  
      We partition the conversation into $K$ contiguous chunks, each
      containing $\lceil N / K\rceil$ turns (the last chunk may be
      shorter).

\item \textbf{Selective token duplication.}  
      Within the \emph{current} chunk we apply the same response-token
      duplication as in Section~\ref{subsec:cust-mask}:
      $r_{i}^{\text{in}}, r_{i}^{\text{out}}$.  
      All earlier chunks act purely as context and therefore retain
      their original, non-duplicated responses.  This progressively
      lowers the number of duplicated tokens as $K$ increases, which is
      the main source of the memory savings reported in
      Table~\ref{tab:kpass}.

\item \textbf{Attention and position IDs.}  
      The custom attention mask and position-ID assignment described in
      Section~\ref{subsec:cust-mask} are applied \emph{only} to the
      duplicated tokens of the active chunk.  Context tokens keep the
      standard causal mask.

\item \textbf{Loss computation.}  
      The label mask is set to \texttt{1} for
      $t_{i}$ and $r_{i}^{\text{out}}$ \emph{inside} the active chunk
      and \texttt{0} elsewhere, so each pass trains only on the new
      turns while reusing earlier content as fixed context.

\end{enumerate}

Conceptually, the $K$-Pass schedule interpolates between the extremes:
\begin{itemize}
    \item $K=N$ reproduces the per-turn baseline
          (no response duplication, minimal memory, maximal passes);
    \item $K=1$ is our 1-Pass method
          (maximum duplication, single pass, fastest).
\end{itemize}

% You can have as much text here as you want. The main body must be at most $8$ pages long.
% For the final version, one more page can be added.
% If you want, you can use an appendix like this one.  

% The $\mathtt{\backslash onecolumn}$ command above can be kept in place if you prefer a one-column appendix, or can be removed if you prefer a two-column appendix.  Apart from this possible change, the style (font size, spacing, margins, page numbering, etc.) should be kept the same as the main body.
%%%%%%%%%%%%%%%%%%%%%%%%%%%%%%%%%%%%%%%%%%%%%%%%%%%%%%%%%%%%%%%%%%%%%%%%%%%%%%%
%%%%%%%%%%%%%%%%%%%%%%%%%%%%%%%%%%%%%%%%%%%%%%%%%%%%%%%%%%%%%%%%%%%%%%%%%%%%%%%

\end{document}